# Defeasible Reasoning in OSCAR

John L. Pollock
Department of Philosophy
University of Arizona
Tucson, Arizona 85721
(e-mail: pollock@arizona.edu)

## 1. General Information

The objective of the OSCAR project is the construction of a general theory of rationality for autonomous agents and its implementation in an AI system. OSCAR is an architecture for rational agents based upon an evolving theory of rational cognition. The general architecture is described in Pollock (1995), and both the implemented system and related papers can be downloaded from http://www.u.arizona.edu/~pollock. The core of OSCAR is a general-purpose defeasible reasoner. OSCAR consists of about 2 Mb of Common LISP code, and runs on any platform.

## 2. Applying the System

*2.1 Methodology*

OSCAR implements a form of defeasible argumentation theory (as described, for example, in Prakken and Vreeswijk 2001). As such, problems are encoded by encoding background knowledge, inputs, and reason-schemas. There is no restriction on the form of encoding, although first-order languages are generally most convenient. OSCAR is equipped with a macro language for the easy encoding of reason-schemas (see section 6).

*2.2 Specifics*

The system is not in any direct sense logic-based. On the other hand, given a set of reason-schemas for first-order logic, OSCAR also becomes a complete theorem prover for first-order logic, and OSCAR is able to reason defeasibly about problems expressed in first-order languages. The style of reasoning in OSCAR is that of a "natural deduction" theorem prover, rather than a more traditional resolution refutation theorem prover.

The motivation for building OSCAR in this way was to provide a platform for defeasible reasoning, but OSCAR turns out to be surprising efficient as a deductive reasoner. In a recent comparison with the highly respected OTTER resolution-refutation theorem prover (http://www.mcs.anl.gov/home/mccume/ar/otter) on a set of 163 problems chosen by Geoff Sutcliffe from the TPTP theorem proving library (Sutcliffe and Suttner 1998), OTTER failed to get 16 while OSCAR failed to get 3. On problems solved by both theorem provers, OSCAR (written in LISP) was on the average 40 times faster than OTTER (written in C).

The principal contribution of OSCAR's defeasible reasoner is that it provides the inference-engine for autonomous rational agents capable of sophisticated reasoning about perception, time, causation, planning, etc. This is illustrated below.

*2.3 Users and Usability*

The potential user need not be an expert in logic or in LISP in order to use the system, because reason-schemas are encoded using the macro language. However, for sophisticated examples, it may be useful to have at least some familiarity with LISP because it is possible to insert bits of LISP code into the reason-schemas. The user must, of course, understand how to reason about the domain being encoded, because it is that reasoning he is encoding. The system is completely general in its application, and has been applied to reasoning about perception, time, causation, the frame problem, classical planning, and is now being used to develop a decision-theoretic planner. I am not sure who is using the system. It is downloaded a lot from my website, and has been used in some advanced AI courses at other universities.

## 3. Evaluating the System

Benchmarks, in the traditional sense, are not appropriate for evaluating a system of this sort. The evaluation of the system is in terms of its ability to encode and solve hard problems in defeasible reasoning. These problems could be called "benchmarks", but no two systems will encode them in the same way. This is illustrated below by the Yale Shooting Problem, which OSCAR encodes and solves very simply. By way of contrast, circumscription-based approaches encode the problem in a completely different (and much more complex) way.

Because there is no uniform encoding for the kinds of problems this system tries to solve, it doesn't make much sense to talk about problem size. OSCAR is able to solve problems that are generally considered difficult, such as Shoham's extended prediction problem (see Pollock 1998) and Stewart Russell's flat tire problem. The system is intended to solve problems of realistic difficulty.

## 4. The OSCAR Architecture

The OSCAR defeasible reasoner provides the inference-engine for the general OSCAR architecture, and is best understood from that perspective. It is natural to distinguish between *epistemic cognition*, which is cognition about what to believe, and *practical cognition*, which is cognition about what to do. The latter is concerned with goal selection, plan construction, plan selection, and plan execution.

In OSCAR, epistemic reasoning is driven by both input from perception and queries passed from practical cognition. This is accomplished in the OSCAR architecture by bidirectional reasoning. The queries passed to epistemic cognition from practical cognition are *epistemic interests*, and OSCAR reasons backwards from them to other epistemic interests. In addition, OSCAR reasons forwards from beliefs to beliefs, until beliefs are produced that discharge the interests.

The principal virtue of OSCAR's epistemic reasoning is that it is capable of performing general-purpose defeasible reasoning. *All sophisticated epistemic cognizers must reason defeasibly.* For example:
- Perception is not always accurate. In order for a cognizer to correct for inaccurate perception, it must draw conclusions defeasibly and be prepared to withdraw them later in the face of conflicting information.
- A sophisticated cognizer must be able to discover general facts about its environment by making particular observations. Obviously, such inductive reasoning must be defeasible because subsequent observations may show the generalization to be false.
- Sophisticated cognizers must reason defeasibly about time, projecting conclusions drawn at one time forwards to future times (temporal projection).
- Certain aspects of planning must be done defeasibly in an autonomous agent operating in a complex environment.

Defeasible reasoning is performed by using defeasible reason-schemas to construct arguments. What makes a reason-schema defeasible is that inferences in accordance with it can be defeated. OSCAR recognizes two kinds of defeaters. *Rebutting defeaters* attack the conclusion of the inference. *Undercutting defeaters* attack the connection between the premise and the conclusion. An undercutting defeater for an inference from *P* to *Q* is a reason for believing it false that *P* would not be true unless *Q* were true. This is symbolized ($P \otimes Q$). More simply, ($P \otimes Q$) can be read "*P* does not guarantee *Q*". For example, something's looking red gives us a defeasible reason for thinking it is red. A reason for thinking it isn't red is a rebutting defeater. However, if we know that it is illuminated by red lights, where that can make something look red when it isn't, that is also defeater but it is not a reason for thinking the object isn't red. Thus it constitutes an undercutting defeater.

Reasoning defeasibly has two parts, (1) constructing arguments for conclusions and (2) deciding what to believe given a set of interacting arguments some of which support defeaters for others. The latter is done by computing defeat statuses and degrees of justification given the set of arguments constructed. OSCAR uses the defeat-status computation described in Pollock (1995). (For comparison with other approaches, see Prakken and Vreeswijk 2001.) This defeat status computation proceeds in terms of the agent's *inference-graph*, which is a data-structure recording the set of arguments thus far constructed. We then define:

> A *partial-status-assignment* for an inference-graph *G* is an assignment of "defeated" and "undefeated" to a subset of the arguments in *G* such that for each argument *A* in *G*:
> 1. if a defeating argument for an inference in *A* is assigned "undefeated", *A* is assigned "defeated";
> 2. if all defeating arguments for inferences in *A* are assigned "defeated", *A* is assigned "undefeated".
>
> A *status-assignment* for an inference-graph *G* is a maximal partial-status-assignment, i.e., a partial-status-assignment not properly contained in any other partial-status-assignment.
>
> An argument *A* is *undefeated* relative to an inference-graph *G* of which it is a member if and only if every status-assignment for *G* assigns "undefeated" to *A*.
>
> A belief is *justified* if and only if it is supported by an argument that is undefeated relative to the inference-graph that represents the agent's current epistemological state.

*Justified beliefs* are those undefeated given the current stage of argument construction. *Warranted conclusions* are those that are undefeated relative to the set of all possible arguments that can be constructed given the current inputs. Raymond Reiter (1980) and David Israel (1980) both observed twenty years ago that when reasoning defeasibly in a rich logical theory like first-order logic, the set of warranted conclusions will not generally be recursively enumerable. This has the consequence that it is impossible to build an automated defeasible reasoner that produces all and only warranted conclusions. In other words, a defeasible reasoner cannot look much like a deductive reasoner. The most we can require is that the reasoner systematically modify its belief set so that it comes to approximate the set of warranted conclusions more and more closely. More precisely, the rules for reasoning should be such that:
> (1) if a proposition *P* is warranted then the reasoner will eventually reach a stage where *P* is justified and stays justified;
> (2) if a proposition *P* is unwarranted then the reasoner will eventually reach a stage where *P* is unjustified and stays unjustified.

This is possible if the set of warranted conclusions is $\Delta_2$, and it is shown in Pollock (1995) that this is true if the reason-schemas are "well behaved". The result is that if we allow the reasoner to draw conclusions "provisionally" and take them back later when necessary, the reasoner can be constructed so that "in the limit" it produces all and only undefeated conclusions. OSCAR is based upon an implementation of this idea. The result is that OSCAR is the only currently implemented system of defeasible reasoning capable of functioning in a rich logical environment that includes full first-order logic.

A distinction can be made between two senses in which OSCAR's reasoning is defeasible. It is *diachronically defeasible* in the sense that the addition of new information can make a previously warranted proposition unwarranted. This is the sense of defeasibility traditionally studied in nonmonotonic logic. It is also *synchronically defeasible* in the sense that further reasoning, without any new inputs, can make a previously justified conclusion unjustified. Notice that human reasoning is also defeasible in both of these senses. In the face of synchronically defeasibility, a reasoner may never be able to stop reasoning, and accordingly there may never be a guarantee that a defeasibly justified conclusion is warranted. However, an agent must act, and as it cannot wait for reasoning to terminate, it must base its practical decisions on its currently justified beliefs.

## 5. Applying Defeasible Reasoning: Perceptual and Temporal Reasoning

Given the ability to perform general-purpose defeasible reasoning, we can go on to provide an agent with reason-schemas for reasoning about specific subject matters. For example, as discussed in my (1998), among the reason-schemas standardly used by OSCAR are the following:

PERCEPTION
> Having a percept at time *t* with content *P* is a defeasible reason to believe *P*-at-*t*.

PERCEPTUAL-RELIABILITY
> "*R* is true and having a percept with content *P* is not a reliable indicator of *P*'s being true when *R* is true" is an undercutting defeater for PERCEPTION.

TEMPORAL-PROJECTION
> "*P*-at-*t*" is a defeasible reason for "*P*-at-($t+\Delta t$)", the strength of the reason being a monotonic decreasing function of $\Delta t$.

STATISTICAL-SYLLOGISM
> "*c* is a *B* & prob(*A*/*B*) is high" is a defeasible reason for "*c* is an *A*".

## 6. Applying Defeasible Reasoning: Implementation

Reasoning proceeds in terms of reasons. A great deal of OSCAR's efficiency derives from the fact that OSCAR reasons bidirectionally. *Backwards-reasons* are used in reasoning backwards, and *forwards-reasons* are used in reasoning forwards. Forwards-reasons are data-structures with the following fields:
- reason-name.
- forwards-premises — a list of forwards-premises.
- backwards-premises — a list of backwards-premises.
- reason-conclusion — a formula.
- defeasible-rule — T if the reason is a defeasible reason, NIL otherwise.
- reason-variables — variables used in pattern-matching to find instances of the reason-premises.
- reason-strength — a real number between 0 and 1, or an expression containing some of the reason-variables and evaluating to a number.

*Forwards-premises* are data-structures encoding the following in-

formation:
- fp-formula — a formula.
- fp-kind — :inference, :percept, or :desire (the default is :inference)
- fp-condition — an optional constraint that must be satisfied by an inference-node for it to instantiate this premise.

Similarly, *backwards-premises* are data-structures encoding the following information:
- bp-formula
- bp-kind
- bp-condition

The use of the premise-kind is to check whether the formula from which a forwards inference proceeds represents a desire, percept, or conclusion (potential belief). The contents of percepts, desires, and conclusions are all encoded as formulas, but the inferences that can be made from them depend upon which kind of item they are. For example, we reason quite differently from the desire that *x* be red, the percept of *x*'s being red, and the conclusion that *x* is red.

*Backwards-reasons* will be data-structures encoding the following information:
- reason-name.
- forwards-premises.
- backwards-premises.
- reason-conclusion — a formula.
- reason-variables — variables used in pattern-matching to find instances of the reason-premises.
- strength — a real number between 0 and 1, or an expression containing some of the reason-variables and evaluating to a number.
- defeasible-rule — T if the reason is a defeasible reason, NIL otherwise.
- reason-condition — a condition that must be satisfied by an interest before the reason is deployed.

*Simple forwards-reasons* have no backwards-premises, and *simple backwards-reasons* have no forwards-premises. Given inference-nodes that instantiate the premises of a simple forwards-reason, the reasoner infers the corresponding instance of the conclusion. Similarly, given an interest that instantiates the conclusion of a simple backwards-reason, the reasoner adopts interest in the corresponding instances of the backwards-premises. Given inference-nodes that discharge those interests, an inference is made to the conclusion from those inference-nodes.

In deductive reasoning, with the exception of a rule of reductio ad absurdum, we are unlikely to encounter any but simple forwards- and backwards-reasons. (This is discussed at greater length in Chapter Two of my 1995.) However, the use of backwards-premises in forwards-reasons and the use of forwards-premises in backwards-reasons provides an invaluable form of control over the way reasoning progresses. This is illustrated in my (1998). *Mixed* forwards- and backwards-reasons are those having both forwards- and backwards-premises. Given inference-nodes that instantiate the forwards-premises of a mixed forwards-reason, the reasoner does not immediately infer the conclusion. Instead the reasoner adopts interest in the corresponding instances of the backwards-premises, and an inference is made only when those interests are discharged. Similarly, given an interest instantiating the conclusion of a mixed backwards-reason, interests are not immediately adopted in the backwards-premises. Interests in the backwards-premises are adopted only when inference-nodes are constructed that instantiate the forwards-premises.

There can also be *degenerate backwards-reasons* that have only forwards-premises. In a degenerate backwards-reason, given an interest instantiating the conclusion, the reasoner then becomes "sensitive to" inference-nodes instantiating the forwards-premises, but does not adopt interest in them (and thereby actively search for arguments to establish them). If appropriate inference-nodes are produced by other reasoning, then an inference is made to the conclusion. Degenerate backwards-reasons are thus much like simple forwards-reasons, except that the conclusion is only drawn if there is an interest in it.

Reasons are most easily defined in OSCAR using the macros DEF-FORWARDS-REASON and DEF-BACKWARDS-REASON:

(def-forwards-reason *symbol*
 :forwards-premises *list of formulas optionally interspersed with expressions of the form (:kind ...) or (:condition ...)*
 :backwards-premises *list of formulas optionally interspersed with expressions of the form (:kind ...) or (:condition ...)*
 :conclusion *formula*
 :strength *number or a an expression containing some of the reason-variables and evaluating to a number.*
 :variables *list of symbols*
 :defeasible? *T or NIL (NIL is the default)*

(def-backwards-reason *symbol*
 :conclusion *list of formulas*
 :forwards-premises *list of formulas optionally interspersed with expressions of the form (:kind ...) or (:condition ...)*
 :backwards-premises *list of formulas optionally interspersed with expressions of the form (:kind ...) or (:condition ...)*
 :condition *this is a predicate applied to the binding produced by the target sequent*
 :strength *number or an expression containing some of the reason-variables and evaluating to a number.*
 :variables *list of symbols*
 :defeasible? *T or NIL (NIL is the default)*

Epistemic reasoning begins from contingent information input into the system in the form of percepts. Percepts are encoded as structures with the following fields:
- percept-content—a formula, without temporal reference built in.
- percept-clarity—a number between 0 and 1, indicating how strong a reason the percept provides for the conclusion of a perceptual inference.
- percept-date—a number.

When a new percept is presented to OSCAR, an inference-node of kind :percept is constructed, having a node-formula that is the percept-content of the percept (this includes the percept-date). This inference-node is then inserted into the inference-queue for processing.

Using the tools described above, we can implement PERCEPTION as a simple forwards-reason:

(def-forwards-reason PERCEPTION
 :forwards-premises "(p at time)" (:kind :percept)
 :conclusion "(p at time)"
 :variables p time
 :defeasible? t
 :strength .98)

When giving an account of a species of defeasible reasoning, it is as important to characterize the defeaters for the defeasible reasons as it is to state the reasons themselves. The only obvious undercutting defeater for PERCEPTION is a reliability defeater, which is of a general sort applicable to all defeasible reasons. Reliability defeaters result from observing that the inference from *P* to *Q* is not, under the present circumstances, reliable. This is

formulated by the following reason-schema:

```
(def-backwards-undercutter PERCEPTUAL-RELIABILITY
 :defeatee PERCEPTION
 :forwards-premises
  "((the probability of p given ((I have a percept with content p) & R)) <= s)"
  (:condition (and (s < 0.99) (temporally-projectible R)))
  "(R at time)"
 :variables p time R s
 :defeasible? t)
```

For an explanation of reason-strengths and projectibility, see my (1998).

Similarly, TEMPORAL-PROJECTION is implemented as follows:

```
(def-backwards-reason TEMPORAL-PROJECTION
 :conclusion "(p at time)"
 :condition (and (temporally-projectible p) (numberp time))
 :forwards-premises
  "(p at time0)"
  (:condition (and (time0 < time*) ((time* - time0) <
log(.5)/log(*temporal-decay*))))
 :variables p time0 time
 :defeasible? T
 :strength (- (* 2 (expt *temporal-decay* (- time time0))) 1))
```

To illustrate the use of these reason-schemas, consider the following problem. First, Fred looks red to me. Later, I am informed by Merrill that I am then wearing blue-tinted glasses. Later still, Fred looks blue to me. All along, I know that the probability is not high of Fred being blue given that Fred looks blue to me but I am wearing blue-tinted glasses. What should I conclude about the color of Fred? Intuitively, Fred's looking red gives me a reason for thinking that Fred is red. Being informed by Merrill that I am wearing blue-tinted glasses gives me a reason for thinking I am wearing blue-tinted glasses. Fred's later looking blue gives me a reason for thinking the world has changed and Fred has become blue. However, my knowledge about the blue-tinted glasses defeats the inference to the conclusion that Fred is blue, reinstating the conclusion that Fred is red. OSCAR's reasoning is diagrammed by figure 1, where the "fuzzy" arrows indicate defeat relations. (For more on this see my 1998.)

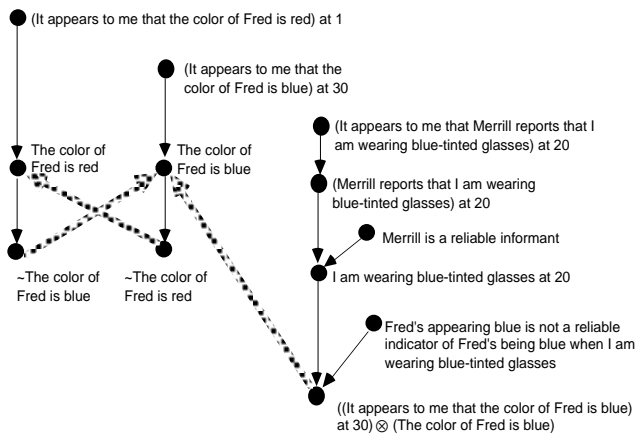

**Figure 1.** Inference graph

## 7. Applying Defeasible Reasoning: Causal Reasoning

For a rational agent to be able to construct plans for making the environment more to its liking, it must be able to reason causally. In particular, it must be able to reason its way through the frame problem. OSCAR implements a solution to the frame problem (see my 1995). It has three constituents:

TEMPORAL-PROJECTION, discussed above.

CAUSAL-IMPLICATION, which allows us to make inferences on the basis of causal knowledge:
    If $t^* > t$, "$A$-at-$t$ and $P$-at-$t$ and ($A$ when $P$ is causally-sufficient for $Q$)" is a defeasible reason for "$Q$-at-$t^*$".

CAUSAL-UNDERCUTTER, which tells us that inferences based on causal knowledge take precedence over inferences based on temporal projection:
    If $t_0 < t < t^*$, "$A$-at-$t$ and $P$-at-$t$ and ($A$ when $P$ is causally-sufficient for $\sim Q$)" is an undercutting defeater for the inference from $Q$-at-$t_0$ to $Q$-at-$t$ by TEMPORAL-PROJECTION.

The latter are implemented as follows:

```
(def-backwards-undercutter CAUSAL-UNDERCUTTER
 :defeatee TEMPORAL-PROJECTION
 :forwards-premises
  "(A when Q is causally sufficient for ~P after an interval interval)"
  "(A at time1)"
  (:condition (and (time0 <= time1) ((time1 + interval) < time)))
 :backwards-premises
  "(Q at time1)"
 :variables A Q P time0 time time* time1 interval op
 :defeasible? T)

(def-backwards-reason CAUSAL-IMPLICATION
 :conclusion "(Q throughout (op time* time**))"
 :condition (and (<= time* time**) ((time** - time*) <
     log(.5)/log(*temporal-decay*)))
 :forwards-premises
 "(A when P is causally sufficient for Q after an interval interval)"
  (:condition (every #'temporally-projectible (conjuncts Q)))
 "(A at time)"
 (:condition
  (or (and (eq op 'clopen) ((time + interval) <= time*) (time* <
time**) ((time** - (time + interval)) < log(.5)/log(*temporal-decay*)))
      (and (eq op 'closed) ((time + interval) < time*) (time* <=
time**) ((time** - (time + interval)) < log(.5)/log(*temporal-decay*)))
      (and (eq op 'open) ((time + interval) <= time*) (time* < time**)
((time** - (time + interval)) < log(.5)/log(*temporal-decay*)))))
 :backwards-premises
  "(P at time)"
 :variables A P Q interval time time* time** op
 :strength (- (* 2 (expt *temporal-decay* (- time** time))) 1)
 :defeasible? T)
```

These principles can be illustrated by applying them to the Yale shooting problem (Hanks and McDermott 1987). I know that the gun being fired while loaded will cause Jones to become dead. I know that the gun is initially loaded, and Jones is initially alive. Later, the gun is fired. Should I conclude that Jones becomes dead? Yes, defeasibly. OSCAR solves this problem by reasoning as in figure 2. By TEMPORAL-PROJECTION, OSCAR has a reason to think that Jones will be alive. By CAUSAL-IMPLICATION, OSCAR has a reason to think that Jones will be dead. By CAUSAL-

UNDERCUTTER, the latter takes precedence, defeating the former (for more details see my 1998). It is shown in my (1998) that quite difficult cases of causal reasoning can be handled equally simply by this system of defeasible reasoning. In particular, there are none of the Baroque reformulations required by the circumscription-based approaches.[1] This, as far as I know, is the first *implemented* fully general solution to the Frame Problem. It is noteworthy how simple it is to implement such principles in OSCAR, making OSCAR a potent tool for epistemological analysis.

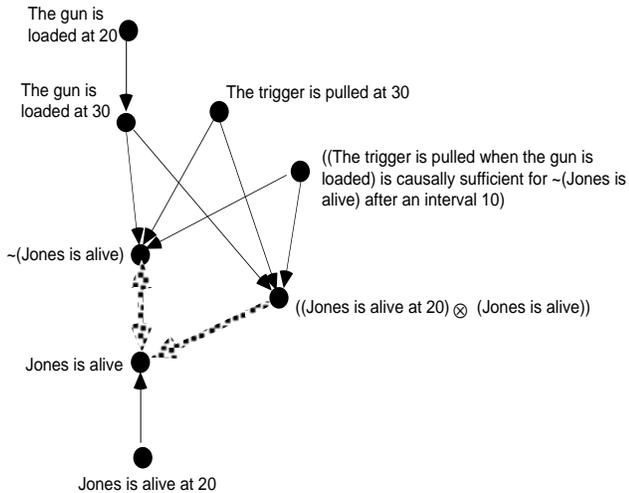

Figure 2. The Yale Shooting Problem

## 8. Applying Defeasible Reasoning: Planning

Given an agent capable of sophisticated epistemic cognition, how can it make use of that in practical cognition? We can regard practical cognition as having four components: goal selection, plan-construction, plan-selection, plan-execution. Although it is natural to think of these as components of practical cognition, most of the work will be carried out by epistemic cognition. To illustrate this, I will focus on plan-construction.

Standard planning algorithms assume that we come to the planning problem with all the knowledge needed to solve it. This assumption fails for autonomous rational agents. The more complex the environment, the more the agent will have to be self-sufficient for knowledge acquisition. The principal function of epistemic cognition is to provide the information needed for practical cognition. As such, the course of epistemic cognition is driven by practical interests. Rather than coming to the planning problem equipped with all the knowledge required for its solution, the planning problem itself directs epistemic cognition, focusing epistemic endeavors on the pursuit of information that will be helpful in solving current planning problems. Paramount among the information an agent may have to acquire in the course of planning is knowledge about the consequences of actions under various circumstances. Sometimes this knowledge can be acquired by reasoning from what is already known. But often it will require empirical investigation. Empirical investigation involves acting, and figuring out what actions to perform requires further planning. So planning drives epistemic investigation, which may in turn drive further planning. In autonomous rational agents operating in a complex environment, planning and epistemic investigation must be interleaved.

---

[1] Examples of circumscription-based approaches are Kartha and Lifschitz (1995), Shanahan (1995), (1996), and (1997).

I assume that rational agents will engage in some form of goal-regression planning. This involves reasoning backwards from goals to subgoals whose achievement will enable an action to achieve a goal. Such reasoning proceeds in terms of causal knowledge of the form "performing action $A$ under circumstances $C$ is causally sufficient for achieving goal $G$". This is symbolized by the *planning-conditional* $(A/C) \Rightarrow G$.

A generally recognized problem for goal-regression planning is that subgoals are typically conjunctions. We usually lack causal knowledge pertaining directly to conjunctions, and must instead use causal knowledge pertaining to the individual conjuncts. We plan separately for the conjuncts of a conjunctive subgoal. When we merge the plans for the conjuncts, we must ensure that the separate plans do not destructively interfere with each other (we must "resolve threats"). Conventional planners assume that the planner already knows the consequences of actions under all circumstances, and so destructive interference can be detected by just checking the consequences. However, an autonomous rational agent may have to engage in arbitrarily much epistemic investigation to detect destructive interference. Even if threats could be detected simply by first-order deductive reasoning, the set of threats would not be recursive. The following theorem is proven in Pollock (1998a):

> If the set of threats is not recursive, then the set of planning ⟨problem,solution⟩ pairs is not recursively enumerable.

> Corollary: A planner that insists upon ruling out threats before merging plans for the conjuncts of a conjunctive goal may never terminate.

If the set of threats is not recursive, a planner must operate defeasibly, *assuming* that there are no threats unless it has reason to believe otherwise. That a plan will achieve a goal is a factual matter, of the sort normally addressed by epistemic cognition. So we can perform plan-search by adopting a set of defeasible reason-schemas for reasoning about plans. The following are examples of such reason-schemas (for details see my 1998a):

GOAL-REGRESSION
  Given an interest in finding a plan for achieving $G$-at-$t$, adopt interest in finding a planning-conditional $(A/C) \Rightarrow G$. Given such a conditional, adopt interest in finding a plan for achieving $C$-at-$t^*$. If it is concluded that a plan *subplan* will achieve $C$-at-$t^*$, construct a plan by (1) adding a new step to the end of *subplan* where the new step prescribes the action $A$-at-$t^*$, (2) adding a constraint $(t^* < t)$ to the ordering-constraints of *subplan*, and (3) adjusting the causal-links appropriately. Infer defeasibly that the new plan will achieve $G$-at-$t$.

SPLIT-CONJUNCTIVE-GOAL
  Given an interest in finding a plan for achieving ($G_1$-at-$t_1$ & $G_2$-at-$t_2$), adopt interest in finding plans $plan_1$ for $G_1$-at-$t_1$ and $plan_2$ for $G_2$-at-$t_2$. Given such plans, infer defeasibly that the result of merging them will achieve ($G_1$-at-$t_1$ & $G_2$-at-$t_2$).

A number of additional reason-schemas are also required, but a complete planner can be constructed in this way. Given a recursive list of consequences for each action, this planner will produce much the same plans as conventional planners like UCPOP (Penberthy and Weld 1992) or PRODIGY (Carbonell et al 1991). But when the list of consequences is nonrecursive the conventional planners will not return solutions, whereas OSCAR will still return the same solutions defeasibly. The search for defeaters may never terminate, but when the agent must act it can do so using its currently justified conclusions, which include the plans that were found defeasibly.

The OSCAR planner is a goal-regression planner, and may seem otiose in light of recent advances in planning stemming from Graphplan (Blum and Furst 1995, 1997) and Satplan (Kautz and Selman 1996, 1998). However, I don't think that is true. First, those planners require complete knowledge of the world, even down to knowing precisely what objects there are in the world and what their properties are. This is not feasible for any realistically complex environment. For example, the current upper limit of the size of the world in which BLACKBOX (Satplan) can operate is reported to be around 53 fluents (Weld 1999). That is impressive from certain perspectives, but very depressing if one wants to do real-world planning. Second, the jury is still out on the question of why these planners work so well. Satplan's strength may lie more in its use of stochastic methods than in anything special about planning (see Brafman, Hoos, and Boutilier 1998). Graphplan's success seems to be based on an interesting feature that it shares with OSCAR, and that may also explain OSCAR's success as a theorem prover. Graphplan controls the exponential plan search by first building a structure (the planning graph) that it can compute in polynomial time, and using that to guide the plan search. Interestingly, on many problems more time is spent on building this guide structure than in the exponentially difficult search for plans. OSCAR works somewhat similarly. OSCAR performs bidirectional search, but the backward search, which builds the interest-graph, proceeds in polynomial time. This is for the same reason Graphplan builds the planning graph in polynomial time. It builds a graph whose nodes can have multiple origins, and it does not try to keep track of a consistent chain of origins as it proceeds. Another way to put this is that the backwards search does not produce "partial proofs" (in the sense that UCPOP produces partial plans with open conditions). The process of using the interest graph to construct plans is still an exponentially difficult process, but like Graphplan, it is common for OSCAR to spend as much time on the construction of the interest graph as on the search for plans. More research is required to see just why this general approach is so effective in speeding up both planning and proof search, and to verify that the same phenomenon is occurring in both cases, but this is at least a plausible conjecture. It suggests that the OSCAR planner should also be able to solve some of the hard problems that Graphplan can solve but UCPOP cannot. Preliminary results suggest that this is true. For example, the OSCAR planner solves Stewart Russell's infamous "flat tire problem" very easily (with an effective branching factor of only 1.02). Graphplan was apparently the first planner to do that successfully. UCPOP cannot solve it. I attribute OSCAR's success on this problem to the fact that OSCAR's approach to solving planning problems is much more like what people do—people also find the flat tire problem easy, which is one of the reasons it has been so puzzling that many planners have difficulty with it. The question remains how successful OSCAR will be on a wide range of planning problems. To get a better answer to this question, OSCAR is being entered in the AIPS2000 planning competition.